# Improved Heterogeneous Distance Functions

**D. Randall Wilson**                                    RANDY@AXON.CS.BYU.EDU
**Tony R. Martinez**                                     MARTINEZ@CS.BYU.EDU
*Computer Science Department*
*Brigham Young University*
*Provo, UT 84602, USA*

## Abstract

Instance-based learning techniques typically handle continuous and linear input values well, but often do not handle nominal input attributes appropriately. The Value Difference Metric (VDM) was designed to find reasonable distance values between nominal attribute values, but it largely ignores continuous attributes, requiring discretization to map continuous values into nominal values. This paper proposes three new heterogeneous distance functions, called the Heterogeneous Value Difference Metric (HVDM), the Interpolated Value Difference Metric (IVDM), and the Windowed Value Difference Metric (WVDM). These new distance functions are designed to handle applications with nominal attributes, continuous attributes, or both. In experiments on 48 applications the new distance metrics achieve higher classification accuracy on average than three previous distance functions on those datasets that have both nominal and continuous attributes.

## 1. Introduction

Instance-Based Learning (IBL) (Aha, Kibler & Albert, 1991; Aha, 1992; Wilson & Martinez, 1993; Wettschereck, Aha & Mohri, 1995; Domingos, 1995) is a paradigm of learning in which algorithms typically store some or all of the $n$ available training examples (*instances*) from a *training set*, $T$, during learning. Each instance has an *input vector* $x$, and an *output class c*. During *generalization*, these systems use a distance function to determine how close a new input vector $y$ is to each stored instance, and use the nearest instance or instances to predict the output class of $y$ (i.e., to *classify* $y$). Some instance-based learning algorithms are referred to as nearest neighbor techniques (Cover & Hart, 1967; Hart, 1968; Dasarathy, 1991), and memory-based reasoning methods (Stanfill & Waltz, 1986; Cost & Salzberg, 1993; Rachlin et al., 1994) overlap significantly with the instance-based paradigm as well. Such algorithms have had much success on a wide variety of *applications* (real-world classification tasks).

Many neural network models also make use of distance functions, including radial basis function networks (Broomhead & Lowe, 1988; Renals & Rohwer, 1989; Wasserman, 1993), counterpropagation networks (Hecht-Nielsen, 1987), ART (Carpenter & Grossberg, 1987), self-organizing maps (Kohonen, 1990) and competitive learning (Rumelhart & McClelland, 1986). Distance functions are also used in many fields besides machine learning and neural networks, including statistics (Atkeson, Moore & Schaal, 1996), pattern recognition (Diday, 1974; Michalski, Stepp & Diday, 1981), and cognitive psychology (Tversky, 1977; Nosofsky, 1986).

There are many distance functions that have been proposed to decide which instance is closest to a given input vector (Michalski, Stepp & Diday, 1981; Diday, 1974). Many of these metrics work well for numerical attributes but do not appropriately handle nominal (i.e., discrete, and perhaps unordered) attributes.

The Value Difference Metric (VDM) (Stanfill & Waltz, 1986) was introduced to define an appropriate distance function for nominal (also called *symbolic*) attributes. The Modified Value Difference Metric (MVDM) uses a different weighting scheme than VDM and is used in the PEBLS system (Cost & Salzberg, 1993; Rachlin et al., 1994). These distance metrics work well in many nominal domains, but they do not handle continuous attributes directly. Instead, they rely upon *discretization* (Lebowitz, 1985; Schlimmer, 1987), which can degrade generalization accuracy (Ventura & Martinez, 1995).

Many real-world applications have both nominal and linear attributes, including, for example, over half of the datasets in the UCI Machine Learning Database Repository (Merz & Murphy, 1996). This paper introduces three new distance functions that are more appropriate than previous functions for applications with both nominal and continuous attributes. These new distance functions can be incorporated into many of the above learning systems and areas of study, and can be augmented with weighting schemes (Wettschereck, Aha & Mohri, 1995; Atkeson, Moore & Schaal, 1996) and other enhancements that each system provides.

The choice of distance function influences the *bias* of a learning algorithm. A bias is "a rule or method that causes an algorithm to choose one generalized output over another" (Mitchell, 1980). A learning algorithm must have a bias in order to generalize, and it has been shown that no learning algorithm can generalize more accurately than any other when summed over all possible problems (Schaffer, 1994) (unless information about the problem *other than the training data* is available). It follows then that no distance function can be strictly better than any other in terms of generalization ability, when considering all possible problems with equal probability.

However, when there is a higher probability of one class of problems occurring than another, some learning algorithms can generalize more accurately than others (Wolpert, 1993). This is not because they are better when summed over all problems, but because the problems on which they perform well are more likely to occur. In this sense, one algorithm or distance function can be an improvement over another in that it has a higher probability of good generalization than another, because it is better matched to the kinds of problems that will likely occur.

Many learning algorithms use a *bias of simplicity* (Mitchell, 1980; Wolpert, 1993) to generalize, and this bias is *appropriate*—meaning that it leads to good generalization accuracy—for a wide variety of real-world applications, though the meaning of *simplicity* varies depending upon the representational language of each learning algorithm. Other biases, such as decisions made on the basis of additional domain knowledge for a particular problem (Mitchell, 1980), can also improve generalization.

In this light, the distance functions presented in this paper are more appropriate than those used for comparison in that they on average yield improved generalization accuracy on a collection of 48 applications. The results are theoretically limited to this set of datasets, but the hope is that these datasets are representative of other problems that will be of interest (and occur frequently) in the real world, and that the distance functions presented here will be useful in such cases, especially those involving both continuous and nominal input attributes.

Section 2 provides background information on distance functions used previously. Section 3

introduces a distance function that combines Euclidean distance and VDM to handle both continuous and nominal attributes. Sections 4 and 5 present two extensions of the Value Difference Metric which allow for direct use of continuous attributes. Section 4 introduces the Interpolated Value Difference Metric (IVDM), which uses interpolation of probabilities to avoid problems related to discretization. Section 5 presents the Windowed Value Difference Metric (WVDM), which uses a more detailed probability density function for a similar interpolation process.

Section 6 presents empirical results comparing three commonly-used distance functions with the three new functions presented in this paper. The results are obtained from using each of the distance functions in an instance-based learning system on 48 datasets. The results indicate that the new heterogeneous distance functions are more appropriate than previously used functions on datasets with both nominal and linear attributes, in that they achieve higher average generalization accuracy on these datasets. Section 7 discusses related work, and Section 8 provides conclusions and future research directions.

## 2. Previous Distance Functions

As mentioned in the introduction, there are many learning systems that depend upon a good distance function to be successful. A variety of distance functions are available for such uses, including the Minkowsky (Batchelor, 1978), Mahalanobis (Nadler & Smith, 1993), Camberra, Chebychev, Quadratic, Correlation, and Chi-square distance metrics (Michalski, Stepp & Diday, 1981; Diday, 1974); the Context-Similarity measure (Biberman, 1994); the Contrast Model (Tversky, 1977); hyperrectangle distance functions (Salzberg, 1991; Domingos, 1995) and others. Several of these functions are defined in Figure 1.

Although there have been many distance functions proposed, by far the most commonly used is the Euclidean Distance function, which is defined as:

$$E(\boldsymbol{x}, \boldsymbol{y}) = \sqrt{\sum_{a=1}^{m} (x_a - y_a)^2} \qquad (1)$$

where $\boldsymbol{x}$ and $\boldsymbol{y}$ are two input vectors (one typically being from a stored instance, and the other an input vector to be classified) and $m$ is the number of input variables (*attributes*) in the application. The square root is often not computed in practice, because the closest instance(s) will still be the closest, regardless of whether the square root is taken.

An alternative function, the *city-block* or *Manhattan* distance function, requires less computation and is defined as:

$$M(\boldsymbol{x}, \boldsymbol{y}) = \sum_{a=1}^{m} |x_a - y_a| \qquad (2)$$

The Euclidean and Manhattan distance functions are equivalent to the Minkowskian $r$-distance function (Batchelor, 1978) with $r = 2$ and 1, respectively.

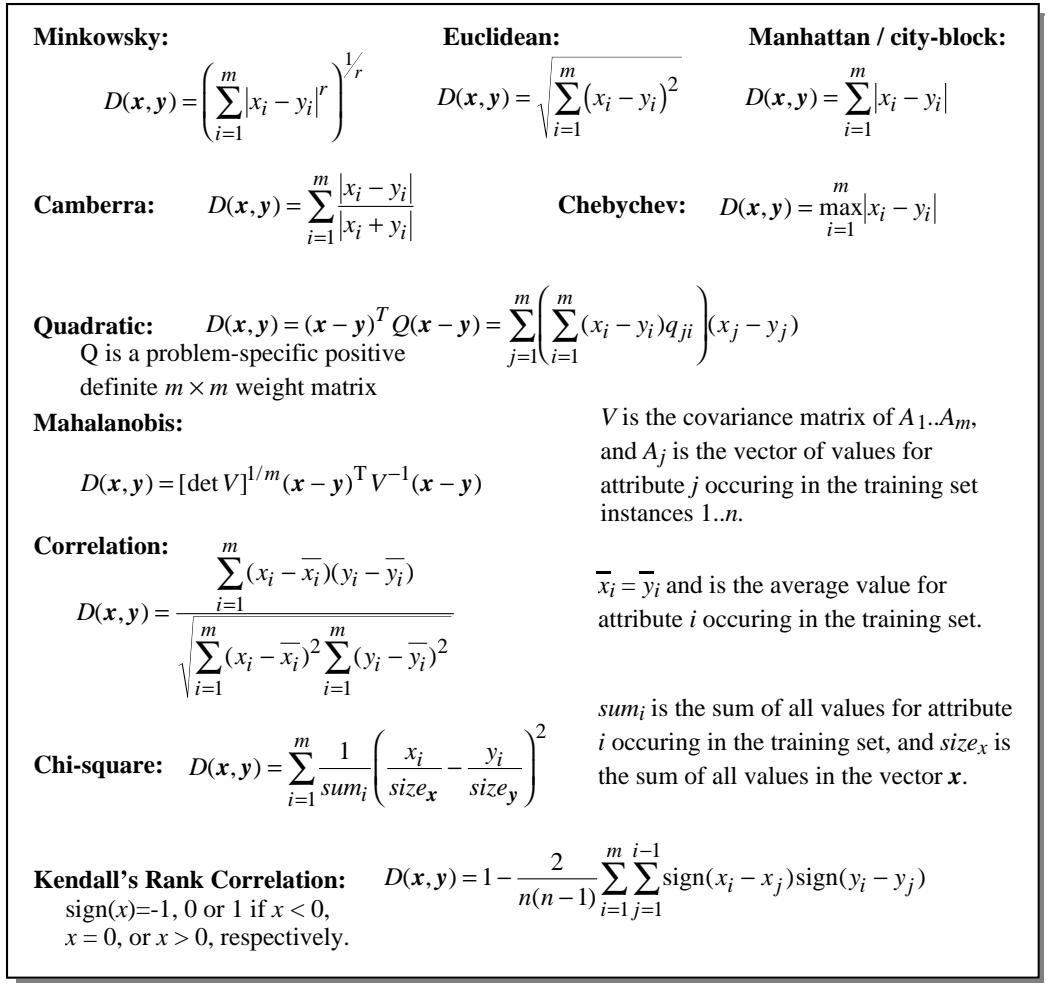

**Minkowsky:** $\quad D(\boldsymbol{x}, \boldsymbol{y}) = \left( \sum_{i=1}^{m} |x_i - y_i|^r \right)^{1/r}$

**Euclidean:** $\quad D(\boldsymbol{x}, \boldsymbol{y}) = \sqrt{\sum_{i=1}^{m} (x_i - y_i)^2}$

**Manhattan / city-block:** $\quad D(\boldsymbol{x}, \boldsymbol{y}) = \sum_{i=1}^{m} |x_i - y_i|$

**Camberra:** $\quad D(\boldsymbol{x}, \boldsymbol{y}) = \sum_{i=1}^{m} \frac{|x_i - y_i|}{|x_i + y_i|}$

**Chebychev:** $\quad D(\boldsymbol{x}, \boldsymbol{y}) = \max_{i=1}^{m} |x_i - y_i|$

**Quadratic:** $\quad D(\boldsymbol{x}, \boldsymbol{y}) = (\boldsymbol{x} - \boldsymbol{y})^T Q (\boldsymbol{x} - \boldsymbol{y}) = \sum_{j=1}^{m} \left( \sum_{i=1}^{m} (x_i - y_i) q_{ji} \right) (x_j - y_j)$

Q is a problem-specific positive definite $m \times m$ weight matrix

**Mahalanobis:**

$D(\boldsymbol{x}, \boldsymbol{y}) = [\det V]^{1/m} (\boldsymbol{x} - \boldsymbol{y})^{\mathrm{T}} V^{-1} (\boldsymbol{x} - \boldsymbol{y})$

$V$ is the covariance matrix of $A_1..A_m$, and $A_j$ is the vector of values for attribute $j$ occuring in the training set instances $1..n$.

**Correlation:**

$D(\boldsymbol{x}, \boldsymbol{y}) = \dfrac{\sum_{i=1}^{m} (x_i - \overline{x_i})(y_i - \overline{y_i})}{\sqrt{\sum_{i=1}^{m} (x_i - \overline{x_i})^2 \sum_{i=1}^{m} (y_i - \overline{y_i})^2}}$

$\overline{x_i} = \overline{y_i}$ and is the average value for attribute $i$ occuring in the training set.

**Chi-square:** $\quad D(\boldsymbol{x}, \boldsymbol{y}) = \sum_{i=1}^{m} \frac{1}{sum_i} \left( \frac{x_i}{size_{\boldsymbol{x}}} - \frac{y_i}{size_{\boldsymbol{y}}} \right)^2$

$sum_i$ is the sum of all values for attribute $i$ occuring in the training set, and $size_{\boldsymbol{x}}$ is the sum of all values in the vector $\boldsymbol{x}$.

**Kendall's Rank Correlation:** $\quad D(\boldsymbol{x}, \boldsymbol{y}) = 1 - \frac{2}{n(n-1)} \sum_{i=1}^{m} \sum_{j=1}^{i-1} \text{sign}(x_i - x_j)\text{sign}(y_i - y_j)$

sign($x$)=-1, 0 or 1 if $x < 0$, $x = 0$, or $x > 0$, respectively.

Figure 1. Equations of selected distance functions.
($\boldsymbol{x}$ and $\boldsymbol{y}$ are vectors of $m$ attribute values).

## 2.1. Normalization

One weakness of the basic Euclidean distance function is that if one of the input attributes has a relatively large range, then it can overpower the other attributes. For example, if an application has just two attributes, *A* and *B*, and *A* can have values from 1 to 1000, and *B* has values only from 1 to 10, then *B*'s influence on the distance function will usually be overpowered by *A*'s influence. Therefore, distances are often *normalized* by dividing the distance for each attribute by the *range* (i.e., maximum-minimum) of that attribute, so that the distance for each attribute is in the approximate range 0..1. In order to avoid outliers, it is also common to divide by the standard deviation instead of range, or to "trim" the range by removing the highest and lowest few percent (e.g., 5%) of the data from consideration in defining the range. It is also possible to map any value outside this range to the minimum or maximum value to avoid normalized values outside the range 0..1. Domain knowledge can often be used to decide which method is most appropriate.

Related to the idea of normalization is that of using attribute weights and other weighting schemes. Many learning systems that use distance functions incorporate various weighting schemes into their distance calculations (Wettschereck, Aha & Mohri, 1995; Atkeson, Moore & Schaal, 1996). The improvements presented in this paper are independent of such schemes, and most of the various weighting schemes (as well as other enhancements such as instance pruning techniques) can be used in conjunction with the new distance functions presented here.

## 2.2. Attribute Types

None of the distance functions shown in Figure 1, including Euclidean distance, appropriately handle non-continuous input attributes.

An attribute can be linear or nominal, and a linear attribute can be continuous or discrete. A *continuous* (or *continuously-valued*) attribute uses real values, such as the mass of a planet or the velocity of an object. A *linear discrete* (or *integer*) attribute can have only a discrete set of linear values, such as *number of children*.

It can be argued that any value stored in a computer is discrete at some level. The reason continuous attributes are treated differently is that they can have so many different values that each value may appear only rarely (perhaps only once in a particular application). This causes problems for algorithms such as VDM (described in Section 2.4) that depend on testing two values for equality, because two continuous values will rarely be equal, though they may be quite close to each other.

A *nominal* (or *symbolic*) attribute is a discrete attribute whose values are not necessarily in any linear order. For example, a variable representing color might have values such as *red, green, blue, brown, black* and *white*, which could be represented by the integers 1 through 6, respectively. Using a linear distance measurement such as (1) or (2) on such values makes little sense in this case.

## 2.3. Heterogeneous Euclidean-Overlap Metric (HEOM)

One way to handle applications with both continuous and nominal attributes is to use a heterogeneous distance function that uses different attribute distance functions on different kinds of attributes. One approach that has been used is to use the *overlap* metric for nominal attributes and normalized Euclidean distance for linear attributes.

For the purposes of comparison during testing, we define a heterogeneous distance function that is similar to that used by IB1, IB2 and IB3 (Aha, Kibler & Albert, 1991; Aha, 1992) as well as that used by Giraud-Carrier & Martinez (1995). This function defines the distance between two values $x$ and $y$ of a given attribute $a$ as:

$$d_a(x,y) = \begin{cases} 1, & \text{if } x \text{ or } y \text{ is unknown, else} \\ overlap(x,y), & \text{if } a \text{ is nominal, else} \\ rn\_diff_a(x,y) \end{cases} \quad (3)$$

Unknown attribute values are handled by returning an attribute distance of 1 (i.e., a maximal distance) if either of the attribute values is unknown. The function *overlap* and the range-normalized difference *rn_diff* are defined as:

$$overlap(x,y) = \begin{cases} 0, & \text{if } x = y \\ 1, & \text{otherwise} \end{cases} \quad (4)$$

$$rn\_diff_a(x,y) = \frac{|x-y|}{range_a} \tag{5}$$

The value $range_a$ is used to normalize the attributes, and is defined as:

$$range_a = max_a - min_a \tag{6}$$

where $max_a$ and $min_a$ are the maximum and minimum values, respectively, observed in the training set for attribute $a$. This means that it is possible for a new input vector to have a value outside this range and produce a difference value greater than one. However, such cases are rare, and when they do occur, a large difference may be acceptable anyway. The normalization serves to scale the attribute down to the point where differences are almost always less than one.

The above definition for $d_a$ returns a value which is (typically) in the range 0..1, whether the attribute is nominal or linear. The overall distance between two (possibly heterogeneous) input vectors $\boldsymbol{x}$ and $\boldsymbol{y}$ is given by the Heterogeneous Euclidean-Overlap Metric function $HEOM(\boldsymbol{x},\boldsymbol{y})$:

$$HEOM(\boldsymbol{x},\boldsymbol{y}) = \sqrt{\sum_{a=1}^{m} d_a(x_a, y_a)^2} \tag{7}$$

This distance function removes the effects of the arbitrary ordering of nominal values, but its overly simplistic approach to handling nominal attributes fails to make use of additional information provided by nominal attribute values that can aid in generalization.

## 2.4. Value Difference Metric (VDM)

The Value Difference Metric (VDM) was introduced by Stanfill and Waltz (1986) to provide an appropriate distance function for nominal attributes. A simplified version of the VDM (without the weighting schemes) defines the distance between two values $x$ and $y$ of an attribute $a$ as:

$$vdm_a(x,y) = \sum_{c=1}^{C} \left| \frac{N_{a,x,c}}{N_{a,x}} - \frac{N_{a,y,c}}{N_{a,y}} \right|^q = \sum_{c=1}^{C} \left| P_{a,x,c} - P_{a,y,c} \right|^q \tag{8}$$

where
- $N_{a,x}$ is the number of instances in the training set $T$ that have value $x$ for attribute $a$;
- $N_{a,x,c}$ is the number of instances in $T$ that have value $x$ for attribute $a$ and output class $c$;
- $C$ is the number of output classes in the problem domain;
- $q$ is a constant, usually 1 or 2; and
- $P_{a,x,c}$ is the conditional probability that the output class is $c$ given that attribute $a$ has the value $x$, i.e., P($c \mid x_a$). As can be seen from (8), $P_{a,x,c}$ is defined as:

$$P_{a,x,c} = \frac{N_{a,x,c}}{N_{a,x}} \tag{9}$$

where $N_{a,x}$ is the sum of $N_{a,x,c}$ over all classes, i.e.,

$$N_{a,x} = \sum_{c=1}^{C} N_{a,x,c} \tag{10}$$

and the sum of $P_{a,x,c}$ over all $C$ classes is 1 for a fixed value of $a$ and $x$.

Using the distance measure $vdm_a(x,y)$, two values are considered to be closer if they have more similar classifications (i.e., more similar correlations with the output classes), regardless of what order the values may be given in. In fact, linear discrete attributes can have their values remapped randomly without changing the resultant distance measurements.

For example, if an attribute *color* has three values *red, green* and *blue*, and the application is to identify whether or not an object is an apple, *red* and *green* would be considered closer than *red* and *blue* because the former two both have similar correlations with the output class *apple*.

The original VDM algorithm (Stanfill & Waltz, 1986) makes use of feature weights that are not included in the above equations, and some variants of VDM (Cost & Salzberg, 1993; Rachlin et al., 1994; Domingos, 1995) have used alternate weighting schemes. As discussed earlier, the new distance functions presented in this paper are independent of such schemes and can in most cases make use of similar enhancements.

One problem with the formulas presented above is that they do not define what should be done when a value appears in a new input vector that never appeared in the training set. If attribute $a$ never has value $x$ in any instance in the training set, then $N_{a,x,c}$ for all $c$ will be 0, and $N_{a,x}$ (which is the sum of $N_{a,x,c}$ over all classes) will also be 0. In such cases $P_{a,x,c} = 0/0$, which is undefined. For nominal attributes, there is no way to know what the probability should be for such a value, since there is no inherent ordering to the values. In this paper we assign $P_{a,x,c}$ the default value of 0 in such cases (though it is also possible to let $P_{a,x,c} = 1/C$, where $C$ is the number of output classes, since the sum of $P_{a,x,c}$ for $c = 1..C$ is always 1.0).

If this distance function is used directly on continuous attributes, the values can all potentially be unique, in which case $N_{a,x}$ is 1 for every value $x$, and $N_{a,x,c}$ is 1 for one value of $c$ and 0 for all others for a given value $x$. In addition, new vectors are likely to have unique values, resulting in the division by zero problem above. Even if the value of 0 is substituted for 0/0, the resulting distance measurement is nearly useless.

Even if all values are not unique, there are often enough different values for a continuous attribute that the statistical sample is unreliably small for each value, and the distance measure is still untrustworthy. Because of these problems, it is inappropriate to use the VDM directly on continuous attributes.

## 2.5. Discretization

One approach to the problem of using VDM on continuous attributes is *discretization* (Lebowitz, 1985; Schlimmer, 1987; Ventura, 1995). Some models that have used the VDM or variants of it (Cost & Salzberg, 1993; Rachlin et al., 1994; Mohri & Tanaka, 1994) have discretized continuous attributes into a somewhat arbitrary number of discrete ranges, and then treated these values as nominal (discrete unordered) values. This method has the advantage of generating a large enough statistical sample for each nominal value that the $P$ values have some significance. However, discretization can lose much of the important information available in the continuous values. For example, two values in the same discretized range are considered equal even if they are on opposite ends of the range. Such effects can reduce generalization accuracy (Ventura & Martinez, 1995).

In this paper we propose three new alternatives, which are presented in the following three sections. Section 3 presents a heterogeneous distance function that uses Euclidean distance for linear attributes and VDM for nominal attributes. This method requires careful attention to the problem of normalization so that neither nominal nor linear attributes are regularly given too much weight.

In Sections 4 and 5 we present two distance functions, the Interpolated Value Difference Metric (IVDM) and the Windowed Value Difference Metric (WVDM), which use discretization to collect statistics and determine values of $P_{a,x,c}$ for continuous values occurring in the training set instances, but then retain the continuous values for later use. During generalization, the value of $P_{a,y,c}$ for a continuous value $y$ is interpolated between two other values of $P$, namely, $P_{a,x1,c}$ and $P_{a,x2,c}$, where $x_1 \leq y \leq x_2$. IVDM and WVDM are essentially different techniques for doing a nonparametric probability density estimation (Tapia & Thompson, 1978) to determine the values of $P$ for each class. A generic version of the VDM algorithm, called the *discretized value difference metric* (DVDM) is used for comparisons with the two new algorithms.

## 3. Heterogeneous Value Difference Metric (HVDM)

As discussed in the previous section, the Euclidean distance function is inappropriate for nominal attributes, and VDM is inappropriate for continuous attributes, so neither is sufficient on its own for use on a heterogeneous application, i.e., one with both nominal and continuous attributes.

In this section, we define a heterogeneous distance function *HVDM* that returns the distance between two input vectors $\boldsymbol{x}$ and $\boldsymbol{y}$. It is defined as follows:

$$HVDM(\boldsymbol{x},\boldsymbol{y}) = \sqrt{\sum_{a=1}^{m} d_a^2(x_a, y_a)} \qquad (11)$$

where $m$ is the number of attributes. The function $d_a(x,y)$ returns a distance between the two values $x$ and $y$ for attribute $a$ and is defined as:

$$d_a(x,y) = \begin{cases} 1, & \text{if } x \text{ or } y \text{ is unknown; otherwise...} \\ normalized\_vdm_a(x,y), & \text{if } a \text{ is nominal} \\ normalized\_diff_a(x,y), & \text{if } a \text{ is linear} \end{cases} \qquad (12)$$

The function $d_a(x,y)$ uses one of two functions (defined below in Section 3.1), depending on whether the attribute is nominal or linear. Note that in practice the square root in (11) is not typically performed because the distance is always positive, and the nearest neighbor(s) will still be nearest whether or not the distance is squared. However, there are some models (e.g., distance-weighted $k$-nearest neighbor, Dudani, 1976) that require the square root to be evaluated.

Many applications contain unknown input values which must be handled appropriately in a practical system (Quinlan, 1989). The function $d_a(x,y)$ therefore returns a distance of 1 if either $x$ or $y$ is unknown, as is done by Aha, Kibler & Albert (1991) and Giraud-Carrier & Martinez (1995). Other more complicated methods have been tried (Wilson & Martinez, 1993), but with little effect on accuracy.

The function HVDM is similar to the function HOEM given in Section 2.3, except that it uses VDM instead of an overlap metric for nominal values and it also normalizes differently. It is also similar to the distance function used by RISE 2.0 (Domingos, 1995), but has some important differences noted below in Section 3.2.

Section 3.1 presents three alternatives for normalizing the nominal and linear attributes. Section 3.2 presents experimental results which show that one of these schemes provides better normalization than the other two on a set of several datasets. Section 3.3 gives empirical results comparing HVDM to two commonly-used distance functions.

### 3.1. Normalization

As discussed in Section 2.1, distances are often *normalized* by dividing the distance for each variable by the range of that attribute, so that the distance for each input variable is in the range 0..1. This is the policy used by HEOM in Section 2.3. However, dividing by the range allows outliers (extreme values) to have a profound effect on the contribution of an attribute. For example, if a variable has values which are in the range 0..10 in almost every case but with one exceptional (and possibly erroneous) value of 50, then dividing by the range would almost always result in a value less than 0.2. A more robust alternative in the presence of outliers is to divide the values by the standard deviation to reduce the effect of extreme values on the typical cases.

For the new heterogeneous distance metric HVDM, the situation is more complicated because the nominal and numeric distance values come from different types of measurements: numeric distances are computed from the difference between two linear values, normalized by standard deviation, while nominal attributes are computed from a sum of $C$ differences of probability values (where $C$ is the number of output classes). It is therefore necessary to find a way to scale these two different kinds of measurements into approximately the same range to give each variable a similar influence on the overall distance measurement.

Since 95% of the values in a normal distribution fall within two standard deviations of the mean, the difference between numeric values is divided by 4 standard deviations to scale each value into a range that is usually of width 1. The function *normalized_diff* is therefore defined as shown below in Equation 13:

$$normalized\_diff_a(x,y) = \frac{|x-y|}{4\sigma_a} \tag{13}$$

where $\sigma_a$ is the standard deviation of the numeric values of attribute $a$.

Three alternatives for the function *normalized_vdm* were considered for use in the heterogeneous distance function. These are labeled N1, N2 and N3, and the definitions of each are given below:

$$N1: normalized\_vdm1_a(x,y) = \sum_{c=1}^{C} \left| \frac{N_{a,x,c}}{N_{a,x}} - \frac{N_{a,y,c}}{N_{a,y}} \right| \tag{14}$$

$$N2: normalized\_vdm2_a(x,y) = \sqrt{\sum_{c=1}^{C} \left| \frac{N_{a,x,c}}{N_{a,x}} - \frac{N_{a,y,c}}{N_{a,y}} \right|^2} \tag{15}$$

$$\text{N3: } normalized\_vdm3_a(x,y) = \sqrt{C * \sum_{c=1}^{C} \left| \frac{N_{a,x,c}}{N_{a,x}} - \frac{N_{a,y,c}}{N_{a,y}} \right|^2} \qquad (16)$$

The function N1 is Equation (8) with $q$=1. This is similar to the formula used in PEBLS (Rachlin et al., 1994) and RISE (Domingos, 1995) for nominal attributes.

N2 uses $q$=2, thus squaring the individual differences. This is analogous to using Euclidean distance instead of Manhattan distance. Though slightly more expensive computationally, this formula was hypothesized to be more robust than N1 because it favors having all of the class correlations fairly similar rather than having some very close and some very different. N1 would not be able to distinguish between these two. In practice the square root is not taken, because the individual attribute distances are themselves squared by the HVDM function.

N3 is the function used in Heterogeneous Radial Basis Function Networks (Wilson & Martinez, 1996), where HVDM was first introduced.

### 3.2. Normalization Experiments

In order to determine whether each normalization scheme N1, N2 and N3 gave unfair weight to either nominal or linear attributes, experiments were run on 15 databases from the machine learning database repository at the University of California, Irvine (Merz & Murphy, 1996). All of the datasets for this experiment have at least some nominal and some linear attributes, and thus require a heterogeneous distance function.

In each experiment, five-fold cross validation was used. For each of the five trials, the distance between each instance in the test set and each instance in the training set was computed. When computing the distance for each attribute, the *normalized_diff* function was used for linear attributes, and the *normalized_vdm* function N1, N2, or N3 was used (in each of the three respective experiments) for nominal attributes.

The average distance (i.e., sum of all distances divided by number of comparisons) was computed for each attribute. The average of all the linear attributes for each database was computed and these averages are listed under the heading "avgLin" in Table 1.

| Database | avgLin | N1 avgNom | N2 avgNom | N3 avgNom | #Nom. | #Lin. | #C |
|---|---|---|---|---|---|---|---|
| Anneal | 0.427 | 0.849 | 0.841 | 0.859 | 29 | 9 | 6 |
| Australian | 0.215 | 0.266 | 0.188 | 0.266 | 8 | 6 | 2 |
| Bridges | 0.328 | 0.579 | 0.324 | 0.808 | 7 | 4 | 7 |
| Crx | 0.141 | 0.268 | 0.193 | 0.268 | 9 | 6 | 2 |
| Echocardiogram | 0.113 | 0.487 | 0.344 | 0.487 | 2 | 7 | 2 |
| Flag | 0.188 | 0.372 | 0.195 | 0.552 | 18 | 10 | 8 |
| Heart | 0.268 | 0.323 | 0.228 | 0.323 | 6 | 7 | 2 |
| Heart.Cleveland | 0.271 | 0.345 | 0.195 | 0.434 | 6 | 7 | 5 |
| Heart.Hungarian | 0.382 | 0.417 | 0.347 | 0.557 | 6 | 7 | 5 |
| Heart.Long-Beach-VA | 0.507 | 0.386 | 0.324 | 0.417 | 6 | 7 | 5 |
| Heart.More | 0.360 | 0.440 | 0.340 | 0.503 | 6 | 7 | 5 |
| Heart.Swiss | 0.263 | 0.390 | 0.329 | 0.421 | 6 | 7 | 5 |
| Hepatitis | 0.271 | 0.205 | 0.158 | 0.205 | 13 | 6 | 2 |
| Horse-Colic | 0.444 | 0.407 | 0.386 | 0.407 | 16 | 7 | 2 |
| Soybean-Large | 0.309 | 0.601 | 0.301 | 0.872 | 29 | 6 | 19 |
| **Average** | **0.299** | **0.422** | **0.313** | **0.492** | **11** | **7** | **5** |

Table 1. Average attribute distance for linear and nominal attributes.

The averages of all the nominal attributes for each of the three normalization schemes are listed under the headings "avgNom" in Table 1 as well. The average distance for linear variables is exactly the same regardless of whether N1, N2 or N3 is used, so this average is given only once. Table 1 also lists the number of nominal ("#Nom.") and number of linear ("#Lin.") attributes in each database, along with the number of output classes ("#C").

As can be seen from the overall averages in the first four columns of the last row of Table 1, N2 is closer than N1 or N3. However, it is important to understand the reasons behind this difference in order to know if the normalization scheme N2 will be more robust in general.

Figures 2-4 graphically display the averages shown in Table 1 under the headings N1, N2 and N3, respectively, ordered from left to right by the number of output classes. We hypothesized that as the number of output classes grows, the normalization would get worse for N3 if it was indeed not appropriate to add the scaling factor $C$ to the sum. The length of each line indicates how much difference there is between the average distance for nominal attributes and linear attributes. An ideal normalization scheme would have a difference of zero, and longer lines indicate worse normalization.

As the number of output classes grows, the difference for N3 between the linear distances and the nominal distances grows wider in most cases. N2, on the other hand, seems to remain quite close independent of the number of output classes. Interestingly, N1 does almost as poorly as N3, even though it does not use the scaling factor $C$. Apparently the squaring factor provides for a more well-rounded distance metric on nominal attributes similar to that provided by using Euclidean distance instead of Manhattan distance on linear attributes.

The underlying hypothesis behind performing normalization is that proper normalization will typically improve generalization accuracy. A nearest neighbor classifier (with $k=1$) was implemented using HVDM as the distance metric. The system was tested on the heterogeneous datasets appearing in Table 1 using the three different normalization schemes discussed above, using ten-fold cross-validation (Schaffer, 1993), and the results are summarized in Table 2. All the normalization schemes used the same training sets and test sets for each trial. Bold entries indicate which scheme had the highest accuracy. An asterisk indicates that the difference was greater than 1% over the next highest scheme.

As can be seen from the table, the normalization scheme N2 had the highest accuracy, and N1 was substantially lower than the other two. N2 and N3 each had the highest accuracy for 8 domains. More significantly, N2 was over 1% higher 5 times compared to N1 being over 1% higher on just one dataset. N3 was higher than the other two on just one dataset, and had a lower average accuracy than N2.

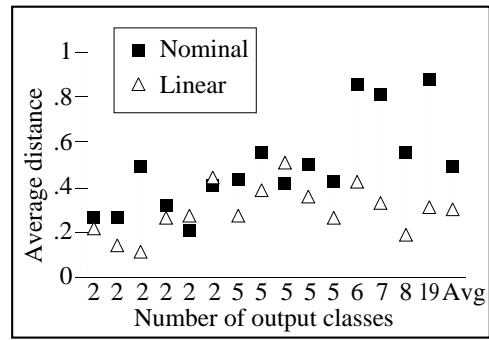

Figure 2. Average distances for N1.

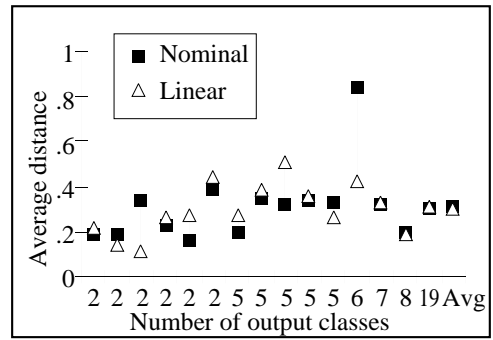

Figure 3. Average distances for N2.

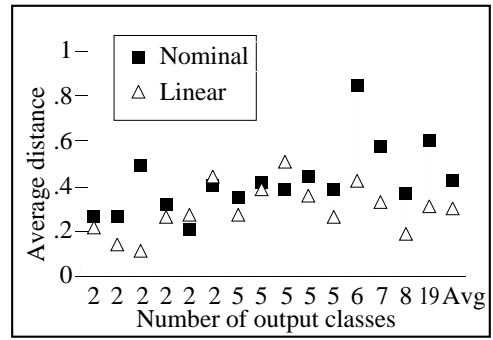

Figure 4. Average distances for N3.

These results support the hypothesis that the normalization scheme N2 achieves higher generalization accuracy than N1 or N3 (on these datasets) due to its more robust normalization though accuracy for N3 is almost as good as N2.

Note that proper normalization will not always necessarily improve generalization accuracy. If one attribute is more important than the others in classification, then giving it a higher weight may improve classification. Therefore, if a more important attribute is given a higher weight accidentally by poor normalization, it may actually improve generalization accuracy. However, this is a random improvement that is not typically the case. Proper normalization should improve generalization in more cases than not when used in typical applications.

| Database | N1 | N2 | N3 |
|---|---|---|---|
| Anneal | 93.98 | 94.61 | **94.99** |
| Australian | 71.30 | 81.45 | **81.59** |
| Bridges | 43.36 | **59.64** | 59.55 |
| Crx | 70.29 | 80.87 | **81.01** |
| Echocardiogram | 70.36 | **94.82** | 94.82 |
| Flag | 28.95 | **55.82*** | 51.50 |
| Heart.Cleveland | 73.88 | **76.56*** | 71.61 |
| Heart.Hungarian | 70.75 | **76.85*** | 75.82 |
| Heart.Long-Beach-Va | 65.50 | 65.50 | **70.00*** |
| Heart.More | 60.03 | 72.09 | **72.48** |
| Heart | 88.46 | **89.49** | **89.49** |
| Heart.Swiss | 74.81 | **78.52*** | 75.19 |
| Hepatitis | 73.50 | 76.67 | **77.33** |
| Horse-Colic | **64.75*** | 60.53 | 60.53 |
| Soybean-Large | 41.45 | **90.88*** | 87.89 |
| **Average** | 66.09 | **76.95** | 76.25 |

Table 2. Generalization accuracy
using N1, N2 and N3.

As a consequence of the above results, N2 is used as the normalization scheme for HVDM, and the function *normalized_vdm* is defined as in (15).

### 3.3. Empirical Results of HVDM vs. Euclidean and HOEM

A nearest neighbor classifier (with *k*=1) using the three distance functions listed in Table 3 was tested on 48 datasets from the UCI machine learning database repository. Of these 48 datasets, the results obtained on the 35 datasets that have at least some nominal attributes are shown in Table 3.

The results are approximately equivalent on datasets with only linear attributes, so the results on the remaining datasets are not shown here, but can be found in Section 6. 10-fold cross-validation was again used, and all three distance metrics used the same training sets and test sets for each trial.

The results of these experiments are shown in Table 3. The first column lists the name of the database ("test" means the database was originally meant to be used as a test set, but was instead used in its entirety as a separate database). The second column shows the results obtained when using the Euclidean distance function normalized by standard deviation on all attributes, including nominal attributes. The next column shows the generalization accuracy obtained by using the HOEM metric, which uses range-normalized Euclidean distance for linear attributes and the overlap metric for nominal attributes. The final column shows the accuracy obtained by using the HVDM distance function which uses the standard-deviation-normalized Euclidean distance (i.e., *normalized_diff* as defined in Equation 13) on linear attributes and the *normalized_vdm* function N2 on nominal attributes.

The highest accuracy obtained for each database is shown in bold. Entries in the Euclid. and HOEM columns that are significantly higher than HVDM (at a 90% or higher confidence level, using a two-tailed paired *t* test) are marked with an asterisk (*). Entries that are significantly lower than HVDM are marked with a "less-than" sign (<).

As can be seen from Table 3, the HVDM distance function's overall average accuracy was higher than that of the other two metrics by over 3%. HVDM achieved as high or higher generalization accuracy than the other two distance functions in 21 of the 35 datasets. The Euclidean distance function was highest in 18 datasets, and HOEM was highest in only 12 datasets.

HVDM was significantly higher than the Euclidean distance function on 10 datasets, and significantly lower on only 3. Similarly, HVDM was higher than HOEM on 6 datasets, and significantly lower on only 4.

These results support the hypothesis that HVDM handles nominal attributes more appropriately than Euclidean distance or the heterogeneous Euclidean-overlap metric, and thus tends to achieve higher generalization accuracy on typical applications.

| Database | Euclid. | HOEM | HVDM |
|---|---|---|---|
| Anneal | **94.99** | 94.61 | 94.61 |
| Audiology | 60.50 < | 72.00 < | **77.50** |
| Audiology.Test | 41.67 < | 75.00 | **78.33** |
| Australian | 80.58 | 81.16 | **81.45** |
| Bridges | 58.64 | 53.73 | **59.64** |
| Crx | 78.99 | **81.01** | 80.87 |
| Echocardiogram | 94.82 | **94.82** | 94.82 |
| Flag | 48.95 < | 48.84 | **55.82** |
| Heart.Cleveland | 73.94 | 74.96 | **76.56** |
| Heart.Hungarian | 73.45 < | 74.47 | **76.85** |
| Heart.Long-Beach-Va | **71.50** | 71.00 * | 65.50 |
| Heart.More | **72.09** | 71.90 | **72.09** |
| Heart.Swiss | **93.53** * | 91.86 | 89.49 |
| Hepatitis | **77.50** | 77.50 | 76.67 |
| Horse-Colic | **65.77** | 60.82 | 60.53 |
| House-Votes-84 | 93.12 < | 93.12 < | **95.17** |
| Image.Segmentation | 92.86 | **93.57** | 92.86 |
| Led+17 | 42.90 < | 42.90 < | **60.70** |
| Led-Creator | **57.20** * | **57.20** * | 56.40 |
| Monks-1.Test | **77.08** | 69.43 | 68.09 |
| Monks-2.Test | 59.04 < | 54.65 < | **97.50** |
| Monks-3.Test | 87.26 < | 78.49 < | **100.00** |
| Mushroom | 100.00 | 100.00 | 100.00 |
| Promoters | 73.73 < | 82.09 < | **92.36** |
| Soybean-Large | 87.26 < | 89.20 | **90.88** |
| Soybean-Small | 100.00 | 100.00 | 100.00 |
| Thyroid.Allbp | 94.89 | 94.89 | **95.00** |
| Thyroid.Allhyper | **97.00** | **97.00** | 96.86 |
| Thyroid.Allhypo | **90.39** | **90.39** * | 90.29 |
| Thyroid.Allrep | **96.14** | **96.14** | 96.11 |
| Thyroid.Dis | **98.21** | **98.21** | 98.21 |
| Thyroid.Hypothyroid | **93.42** | **93.42** | 93.36 |
| Thyroid.Sick-Euthyroid | 68.23 | **68.23** | 68.23 |
| Thyroid.Sick | **86.93** * | 86.89 * | 86.61 |
| Zoo | 97.78 | 94.44 | **98.89** |
| **Average**: | 79.44 | 80.11 | **83.38** |

Table 3. Generalization accuracy of the Euclidean, HOEM, and HVDM distance functions.

## 4. Interpolated Value Difference Metric (IVDM)

In this section and Section 5 we introduce distance functions that allow VDM to be applied directly to continuous attributes. This alleviates the need for normalization between attributes. It also in some cases provides a better measure of distance for continuous attributes than linear distance.

For example, consider an application with an input attribute *height* and an output class that indicates whether a person is a good candidate to be a fighter pilot in a particular airplane. Those individuals with heights significantly below *or* above the preferred height might both be considered poor candidates, and thus it could be beneficial to consider their heights as more similar to each other than to those of the preferred height, even though they are farther apart in a linear sense.

On the other hand, linear attributes for which linearly distant values tend to indicate different classifications should also be handled appropriately. The Interpolated Value Difference Metric (IVDM) handles both of these situations, and handles heterogeneous applications robustly.

A generic version of the VDM distance function, called the *discretized value difference metric* (DVDM) will be used for comparisons with extensions of VDM presented in this paper.

## 4.1. IVDM Learning Algorithm

The original value difference metric (VDM) uses statistics derived from the training set instances to determine a probability $P_{a,x,c}$ that the output class is $c$ given the input value $x$ for attribute $a$.

When using IVDM, continuous values are discretized into $s$ equal-width intervals (though the continuous values are also retained for later use), where $s$ is an integer supplied by the user. Unfortunately, there is currently little guidance on what value of $s$ to use. A value that is too large will reduce the statistical strength of the values of $P$, while a value too small will not allow for discrimination among classes. For the purposes of this paper, we use a heuristic to determine $s$ automatically: let $s$ be 5 or $C$, whichever is greatest, where $C$ is the number of output classes in the problem domain. Current research is examining more sophisticated techniques for determining good values of $s$, such as cross-validation, or other statistical methods (e.g., Tapia & Thompson, 1978, p. 67). (Early experimental results indicate that the value of $s$ may not be critical as long as $s \geq C$ and $s \ll n$, where $n$ is the number of instances in the training set.)

The width $w_a$ of a discretized interval for attribute $a$ is given by:

$$w_a = \frac{|max_a - min_a|}{s} \tag{17}$$

where $max_a$ and $min_a$ are the maximum and minimum value, respectively, occurring in the training set for attribute $a$.

As an example, consider the *Iris* database from the UCI machine learning databases. The *Iris* database has four continuous input attributes, the first of which is *sepal length*. Let $T$ be a training set consisting of 90% of the 150 available training instances, and $S$ be a test set consisting of the remaining 10%.

In one such division of the training set, the values in $T$ for the *sepal length* attribute ranged from 4.3 to 7.9. There are only three output classes in this database, so we let $s$=5, resulting in a width of |7.9 - 4.3| / 5 = 0.72. Note that since the discretization is part of the learning process, it would be unfair to use any instances in the test set to help determine how to discretize the values. The discretized value $v$ of a continuous value $x$ for attribute $a$ is an integer from 1 to $s$, and is given by:

$$v = discretize_a(x) = \begin{cases} x, \text{ if } a \text{ is discrete, else} \\ s, \text{ if } x = max_a, \text{ else} \\ \lfloor (x - min_a) / w_a \rfloor + 1 \end{cases} \tag{18}$$

After deciding upon $s$ and finding $w_a$, the discretized values of continuous attributes can be used just like discrete values of nominal attributes in finding $P_{a,x,c}$. Figure 5 lists pseudo-code for how this is done.

```
LearnP(training set T)
    For each attribute a
        For each instance i in T
            Let x be the input value for attribute a of instance i.
            v = discretize_a(x) [which is just x if a is discrete]
            Let c be the output class of instance i.
            Increment N_{a,v,c} by 1.
            Increment N_{a,v} by 1.
        For each discrete value v (of attribute a)
            For each class c
                If N_{a,v}=0
                    Then P_{a,v,c}=0
                Else P_{a,v,c} = N_{a,v,c} / N_{a,v}
    Return 3-D array P_{a,v,c}.
```

Figure 5. Pseudo code for finding $P_{a,x,c}$.

For the first attribute of the *Iris* database, the values of $P_{a,x,c}$ are displayed in Figure 6. For each of the five discretized ranges of $x$, the probability for each of the three corresponding output classes are shown as the bar heights. Note that the heights of the three bars sum to 1.0 for each discretized range. The bold integers indicate the discretized value of each range. For example, a sepal length greater than or equal to 5.74 but less than 6.46 would have a discretized value of 3.

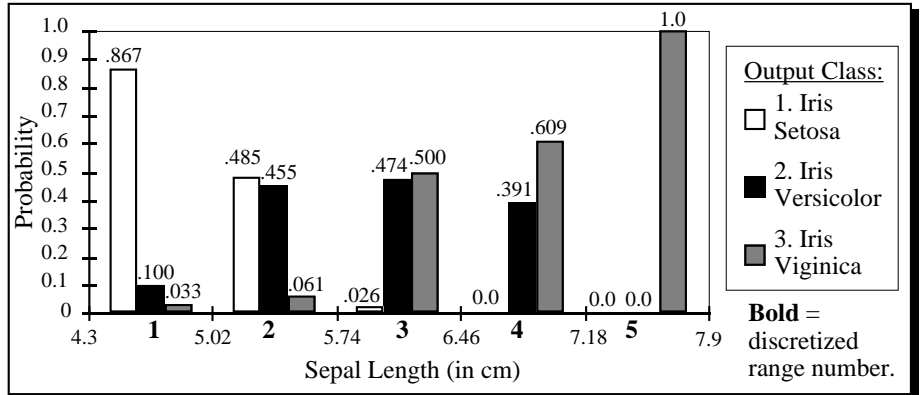

Figure 6. $P_{a,x,c}$ for $a$=1, $x$=1..5, $c$=1..3, on the first attribute of the *Iris* database.

## 4.2. IVDM and DVDM Generalization

Thus far the DVDM and IVDM algorithms learn identically. However, at this point the DVDM algorithm need not retain the original continuous values because it will use only the discretized values during generalization. On the other hand, the IVDM will use the continuous values.

During generalization, an algorithm such as a nearest neighbor classifier can use the distance function DVDM, which is defined as follows:

$$DVDM(\boldsymbol{x},\boldsymbol{y}) = \sum_{a=1}^{m} \left| vdm_a(discretize_a(x_a), discretize_a(y_a)) \right|^2 \qquad (19)$$

where *discretize*$_a$ is as defined in Equation (18) and *vdm*$_a$ is defined as in Equation (8), with $q$=2. We repeat it here for convenience:

$$vdm_a(x,y) = \sum_{c=1}^{C} \left| P_{a,x,c} - P_{a,y,c} \right|^2 \tag{20}$$

Unknown input values (Quinlan, 1989) are treated as simply another discrete value, as was done in (Domingos, 1995).

| | | Input Attributes | | | | |
|---|---|---|---|---|---|---|
| | 1 | 2 | 3 | 4 | | Output Class |
| *A*: | 5.0 | 3.6 | 1.4 | 0.2 | -> | 1 (Iris Setosa) |
| *B*: | 5.7 | 2.8 | 4.5 | 1.3 | -> | 2 (Iris Versicolor) |
| *y*: | 5.1 | 3.8 | 1.9 | 0.4 | | |

Table 4. Example from the *Iris* database.

As an example, consider two training instances *A* and *B* as shown in Table 4, and a new input vector *y* to be classified. For attribute $a$=1, the discretized values for *A*, *B*, and *y* are 1, 2, and 2, respectively. Using values from Figure 6, the distance for attribute 1 between *y* and *A* is:

$$|.867\text{-}.485|^2 + |.1\text{-}.455|^2 + |.033\text{-}.061|^2 = .273$$

while the distance between *y* and *B* is 0, since they have the same discretized value.

Note that *y* and *B* have values on different ends of range 2, and are not actually nearly as close as *y* and *A* are. In spite of this fact, the discretized distance function says that *y* and *B* are equal because they happen to fall into the same discretized range.

IVDM uses interpolation to alleviate such problems. IVDM assumes that the $P_{a,x,c}$ values hold true only at the midpoint of each range, and interpolates between midpoints to find $P$ for other attribute values.

Figure 7 shows the $P$ values for the second output class (*Iris Versicolor*) as a function of the first attribute value (*sepal length*). The dashed line indicates what $P$ value is used by DVDM, and the solid line shows what IVDM uses.

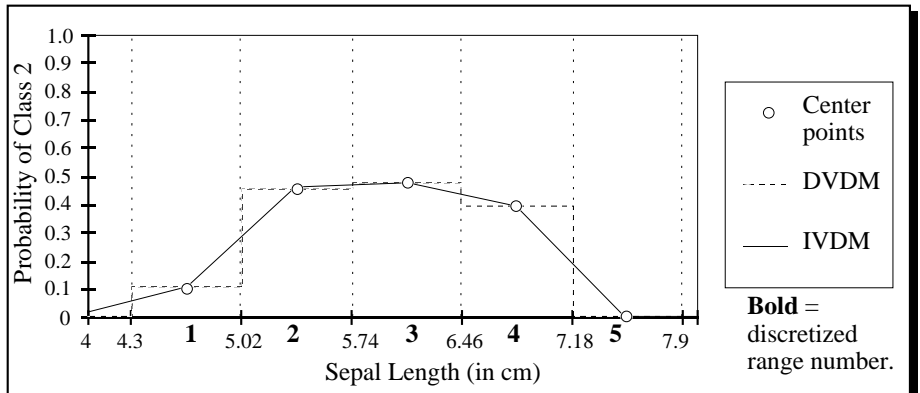

Figure 7. $P_{1,x,2}$ values from the DVDM and IVDM for attribute 1, class 2 of the *Iris* database.

The distance function for the Interpolated Value Difference Metric is defined as:

$$IVDM(\boldsymbol{x}, \boldsymbol{y}) = \sum_{a=1}^{m} ivdm_a(x_a, y_a)^2 \qquad (21)$$

where $ivdm_a$ is defined as:

$$ivdm_a(x, y) = \begin{cases} vdm_a(x, y), & \text{if } a \text{ is discrete} \\ \sum_{c=1}^{C} |p_{a,c}(x) - p_{a,c}(y)|^2, & \text{otherwise} \end{cases} \qquad (22)$$

The formula for determining the interpolated probability value $p_{a,c}(x)$ of a continuous value $x$ for attribute $a$ and class $c$ is:

$$p_{a,c}(x) = P_{a,u,c} + \left( \frac{x - mid_{a,u}}{mid_{a,u+1} - mid_{a,u}} \right) * (P_{a,u+1,c} - P_{a,u,c}) \qquad (23)$$

In this equation, $mid_{a,u}$ and $mid_{a,u+1}$ are midpoints of two consecutive discretized ranges such that $mid_{a,u} \leq x < mid_{a,u+1}$. $P_{a,u,c}$ is the probability value of the discretized range $u$, which is taken to be the probability value of the midpoint of range $u$ (and similarly for $P_{a,u+1,c}$). The value of $u$ is found by first setting $u = discretize_a(x)$, and then subtracting 1 from $u$ if $x < mid_{a,u}$. The value of $mid_{a,u}$ can then be found as follows:

$$mid_{a,u} = min_a + width_a * (u + .5) \qquad (24)$$

Figure 8 shows the values of $p_{a,c}(x)$ for attribute $a=1$ of the *Iris* database for all three output classes (i.e. $c=1$, 2, and 3). Since there are no data points outside the range $min_a..max_a$, the probability value $P_{a,u,c}$ is taken to be 0 when $u < 1$ or $u > s$, which can be seen visually by the diagonal lines sloping toward zero on the outer edges of the graph. Note that the sum of the probabilities for the three output classes sum to 1.0 at every point from the midpoint of range 1 through the midpoint of range 5.

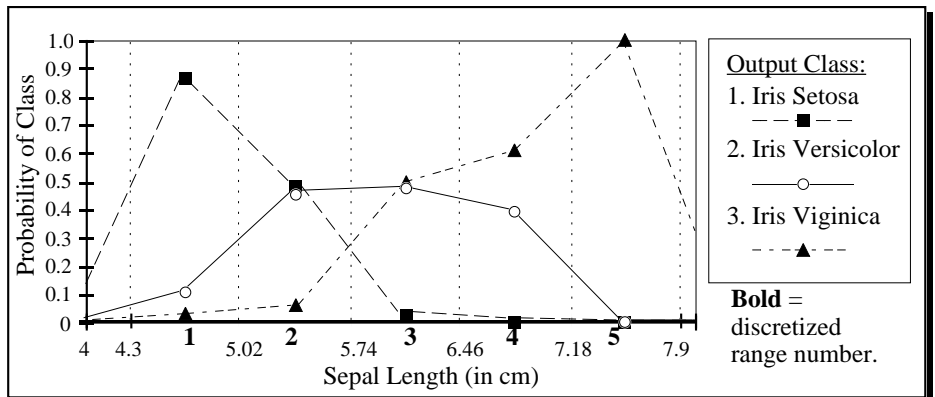

Figure 8. Interpolated probability values for attribute 1 of the *Iris* database.

| | value | $p_{1,1}(v)$ | $p_{1,2}(v)$ | $p_{1,3}(v)$ | $ivdm_1(v,y)$ | $vdm_1(v,y)$ |
|---|---|---|---|---|---|---|
| A | 5.0 | .687 | .268 | .046 | **.005** | .273 |
| B | 5.7 | .281 | .463 | .256 | **.188** | 0 |
| y | 5.1 | .634 | .317 | .050 | | |

Table 5. Example of *ivdm* vs. *vdm*.

Using IVDM on the example instances in Table 4, the values for the first attribute are not discretized as they are with DVDM, but are used to find interpolated probability values. In that example, **y** has a value of 5.1, so $p_{1,c}(x)$ interpolates between midpoints 1 and 2, returning the values shown in Table 5 for each of the three classes. Instance *A* has a value of 5.0, which also falls between midpoints 1 and 2, but instance *B* has a value of 5.7, which falls between midpoints 2 and 3.

As can be seen from Table 5, IVDM (using the single-attribute distance function *ivdm*) returns a distance which indicates that **y** is closer to *A* than *B* (for the first attribute), which is certainly the case here. DVDM (using the discretized *vdm*), on the other hand, returns a distance which indicates that the value of **y** is *equal* to that of *B*, and quite far from *A*, illustrating the problems involved with using discretization.

The IVDM and DVDM algorithms were implemented and tested on 48 datasets from the UCI machine learning databases. The results for the 34 datasets that contain at least some continuous attributes are shown in Table 6. (Since IVDM and DVDM are equivalent on domains with only discrete attributes, the results on the remaining datasets are deferred to Section 6.) 10-fold cross-validation was again used, and the average accuracy for each database over all 10 trials is shown in Table 6. Bold values indicate which value was highest for each dataset. Asterisks (*) indicates that the difference is statistically significant at a 90% confidence level or higher, using a two-tailed paired *t*-test.

On this set of datasets, IVDM had a higher average generalization accuracy overall than the discretized algorithm. IVDM obtained higher generalization accuracy than DVDM in 23 out of 34 cases, 13 of which were significant at the 90% level or above. DVDM had a higher

| Database | DVDM | IVDM |
|---|---|---|
| Annealing | 94.99 | **96.11** * |
| Australian | **83.04** * | 80.58 |
| Bridges | 56.73 | **60.55** |
| Credit Screening | **80.14** | 80.14 |
| Echocardiogram | 100.00 | 100.00 |
| Flag | **58.76** | 57.66 |
| Glass | 56.06 | **70.54** * |
| Heart Disease | 80.37 | **81.85** |
| Heart (Cleveland) | **79.86** | 78.90 |
| Heart (Hungarian) | **81.30** | 80.98 |
| Heart (Long-Beach-Va) | **71.00** | 66.00 |
| Heart (More) | 72.29 | **73.33** |
| Heart (Swiss) | **88.59** | 87.88 |
| Hepatitis | 80.58 | **82.58** |
| Horse-Colic | 76.75 | **76.78** |
| Image Segmentation | 92.38 | **92.86** |
| Ionosphere | **92.60** | 91.17 |
| Iris | 92.00 | **94.67** |
| Liver Disorders | 55.04 | **58.23** |
| Pima Indians Diabetes | **71.89** | 69.28 |
| Satellite Image | 87.06 | **89.79** * |
| Shuttle | 96.17 | **99.77** * |
| Sonar | 78.45 | **84.17** |
| Thyroid (Allbp) | 94.86 | **95.32** |
| Thyroid (Allhyper) | 96.93 | **97.86** * |
| Thyroid (Allhypo) | 89.36 | **96.07** * |
| Thyroid (Allrep) | 96.86 | **98.43** * |
| Thyroid (Dis) | **98.29** | 98.04 |
| Thyroid (Hypothyroid) | 93.01 | **98.07** * |
| Thyroid (Sick) | 88.24 | **95.07** * |
| Thyroid (Sick-Euthyroid) | 88.82 | **96.86** * |
| Vehicle | 63.72 | **69.27** * |
| Vowel | 91.47 | **97.53** * |
| Wine | 94.38 | **97.78** * |
| **Average:** | 83.08 | **85.22** |

Table 6. Generalization for DVDM vs. IVDM.

accuracy in 9 cases, but only one of those had a difference that was statistically significant.

These results indicate that the interpolated distance function is typically more appropriate than the discretized value difference metric for applications with one or more continuous attributes. Section 6 contains further comparisons of IVDM with other distance functions.

## 5. Windowed Value Difference Metric (WVDM)

The IVDM algorithm can be thought of as sampling the value of $P_{a,u,c}$ at the midpoint $mid_{a,u}$ of each discretized range $u$. $P$ is sampled by first finding the instances that have a value for attribute $a$ in the range $mid_{a,u} \pm w_a / 2$. $N_{a,u}$ is incremented once for each such instance, and $N_{a,u,c}$ is also incremented for each instance whose output class is $c$, after which $P_{a,u,c} = N_{a,u,c} / N_{a,u}$ is computed. IVDM then interpolates between these sampled points to provide a continuous but rough approximation to the function $p_{a,c}(x)$. It is possible to sample $P$ at more points and thus provide a closer approximation to the function $p_{a,c}(x)$, which may in turn provide for more accurate distance measurements between values.

Figure 9 shows pseudo-code for the Windowed Value Difference Metric (WVDM). The WVDM samples the value of $P_{a,x,c}$ at each value $x$ occurring in the training set for each

Define:
    *instance*[*a*][1..*n*] as the list of all *n* instances in *T* sorted in ascending order by attribute *a*.
    *instance*[*a*][*i*].*val*[*a*]  as the value of attribute *a* for *instance*[*a*][*i*].
    *x*            as the center value of the current window, i.e., *x*=*instance*[*a*][*i*].*val*[*a*].
    *p*[*a*][*i*][*c*]     as the probability $P_{a,x,c}$ that the output class is *c* given the input value *x* for
                       attribute *a*. Note that *i* is an index, the not value itself.
    *N*[*c*]        as the number $N_{a,x,c}$ of instances in the current window with output class *c*.
    *N*           as the total number $N_{a,x}$ of instances in the current window.
    *instance*[*a*][*in*]   as the first instance in the window.
    *instance*[*a*][*out*]   as the first instance outside the window. (i.e., the window contains
                       instances *instance*[*a*][*in*..*out*-1]).
    *w*[*a*]       as the window width for attribute *a*.

**LearnWVDM**(training set T)
    For each continuous attribute *a*
        Sort *instance*[*a*][1..*n*] in ascending order by attribute *a*, using a quicksort.
        Initialize *N* and *N*[*c*] to 0, and *in* and *out* to 1 (i.e., start with an empty window).
        For each *i*=1..*n*
            Let *x*=*instance*[*a*][*i*].*val*[*a*].
            // Expand window to include all instances in range
            While (*out* < *n*) and (*instance*[*a*][*out*].*val*[*a*] < (*x* + *w*[*a*]/2))
                Increment *N*[*c*], where *c*=the class of *instance*[*a*][*out*].
                Increment *N*.
                Increment *out*.
            // Shrink window to exclude instances no longer in range
            While (*in* < *out*) and (*instance*[*a*][*in*].*val*[*a*] < (*x* - *w*[*a*]/2))
                Decrement *N*[*c*], where *c*=the class of *instance*[*a*][*in*].
                Decrement *N*.
                Increment *in*.
            // Compute the probability value for each class from the current window
            for each class *c*=1..*C*
                *p*[*a*][*i*][*c*] = *N*[*c*] / *N*. (i.e., $P_{a,x,c} = N_{a,x,c} / N_{a,x}$).
        Return the 3-D array *p*[*a*][*i*][*c*].

Figure 9. Pseudo code for the WVDM learning algorithm.

attribute $a$, instead of only at the midpoints of each range. In fact, the discretized ranges are not even used by WVDM on continuous attributes, except to determine an appropriate *window width*, $w_a$, which is the same as the range width used in DVDM and IVDM. The pseudo-code for the learning algorithm used to determine $P_{a,x,c}$ for each attribute value $x$ is given in Figure 9.

For each value $x$ occurring in the training set for attribute $a$, $P$ is sampled by finding the instances that have a value for attribute $a$ in the range $x \pm w_a / 2$, and then computing $N_{a,x}$, $N_{a,x,c}$, and $P_{a,x,c} = N_{a,x,c} / N_{a,x}$ as before. Thus, instead of having a fixed number $s$ of sampling points, a *window* of instances, centered on each training instance, is used for determining the probability at a given point. This technique is similar in concept to *shifted histogram estimators* (Rosenblatt, 1956) and to *Parzen window* techniques (Parzen, 1962).

For each attribute the values are sorted (using an O($n$log$n$) sorting algorithm) so as to allow a sliding window to be used and thus collect the needed statistics in O($n$) time for each attribute. The sorted order is retained for each attribute so that a binary search can be performed in O(log $n$) time during generalization.

Values occurring between the sampled points are interpolated just as in IVDM, except that there are now many more points available, so a new value will be interpolated between two closer, more precise values than with IVDM.

---

**WVDM_Find_P**(attribute $a$,continuous value $x$)
    // Find $P_{a,x,c}$ for $c=1..C$, given a value $x$ for attribute $a$.
    Find $i$ such that *instance*[$a$][$i$].*val*[$a$] $\leq x \leq$ *instance*[$a$][$i$+1].*val*[$a$] (binary search).
    $x_1 = $ *instance*[$a$][$i$].*val*[$a$]        (unless $i$<1, in which case $x_1 = min[a] - (w[a] / 2)$)
    $x_2 = $ *instance*[$a$][$i$+1].*val*[$a$]    (unless $i$>$n$, in which case $x_2 = max[a] + (w[a] / 2)$)
    For each class $c=1..C$
        $p_1 = p[a][i][c]$        (unless $i$<1, in which case $p_1$=0)
        $p_2 = p[a][i+1][c]$    (unless $i$>$n$, in which case $p_2$=0)
        $P_{a,x,c} = p_1 + ((x-x_1)/(x_2-x_1)) * (p_2 - p_1)$
    Return array $P_{a,x,1..C}$.

---

Figure 10. Pseudo-code for the WVDM probability interpolation (see Figure 9 for definitions).

The pseudo-code for the interpolation algorithm is given in Figure 10. This algorithm takes a value $x$ for attribute $a$ and returns a vector of $C$ probability values $P_{a,x,c}$ for $c=1..C$. It first does a binary search to find the two consecutive instances in the sorted list of instances for attribute $a$ that surround $x$. The probability for each class is then interpolated between that stored for each of these two surrounding instances. (The exceptions noted in parenthesis handle outlying values by interpolating towards 0 as is done in IVDM.)

Once the probability values for each of an input vector's attribute values are computed, they can be used in the *vdm* function just as the discrete probability values are.

The WVDM distance function is defined as:

$$WVDM(\boldsymbol{x}, \boldsymbol{y}) = \sum_{a=1}^{m} wvdm_a(x_a, y_a)^2 \tag{25}$$

and *wvdm$_a$* is defined as:

$$wvdm_a(x,y) = \begin{cases} vdm_a(x,y), & \text{if } a \text{ is discrete} \\ \sum_{c=1}^{C} \left| P_{a,x,c} - P_{a,y,c} \right|^2, & \text{otherwise} \end{cases} \tag{26}$$

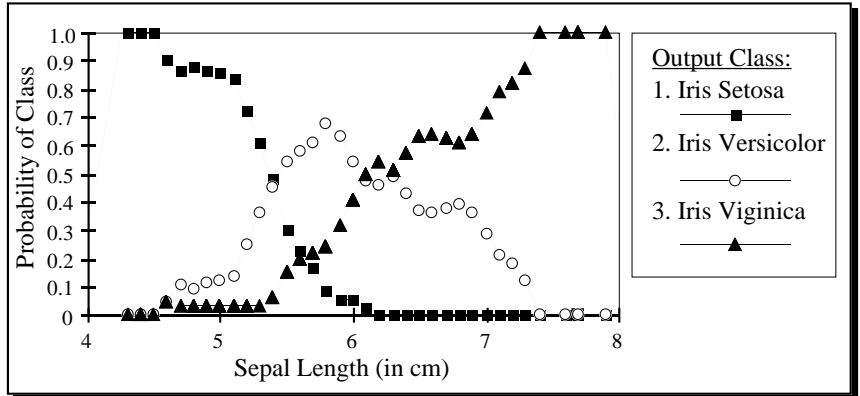

Figure 11. Example of the WVDM probability landscape.

where $P_{a,x,c}$ is the interpolated probability value for the continuous value $x$ as computed in Figure 10. Note that we are typically finding the distance between a new input vector and an instance in the training set. Since the instances in the training set were used to define the probability at each of their attribute values, the binary search and interpolation is unnecessary for training instances because they can immediately recall their stored probability values, unless pruning techniques have been used.

One drawback to this approach is the increased storage needed to retain $C$ probability values for each attribute value in the training set. Execution time is not significantly increased over IVDM or DVDM. (See Section 6.2 for a discussion on efficiency considerations).

Figure 11 shows the probability values for each of the three classes for the first attribute of the *Iris* database again, this time using the windowed sampling technique. Comparing Figure 11 with Figure 8 reveals that on this attribute IVDM provides approximately the same overall shape, but misses much of the detail. For example, the peak occurring for output class 2 at approximately *sepal length*=5.75. In Figure 8 there is a flat line which misses this peak entirely, due mostly to the somewhat arbitrary position of the midpoints at which the probability values are sampled.

Table 7 summarizes the results of testing the WVDM algorithm on the same datasets as DVDM and IVDM. A bold entry again indicates the highest of the two accuracy measurements, and an asterisk (*) indicates a difference that is statistically significant at the 90% confidence level, using a two-tailed paired *t*-test.

| Database | DVDM | WVDM |
|---|---|---|
| Annealing | 94.99 | **95.87** |
| Australian | **83.04** | 82.46 |
| Bridges | **56.73** | 56.64 |
| Credit Screening | 80.14 | **81.45** |
| Echocardiogram | **100.00** | 98.57 |
| Flag | **58.76** | 58.74 |
| Glass | 56.06 | **71.49** * |
| Heart Disease | 80.37 | **82.96** |
| Heart (Cleveland) | 79.86 | **80.23** |
| Heart (Hungarian) | **81.30** | 79.26 |
| Heart (Long-Beach-Va) | **71.00** | 68.00 |
| Heart (More) | 72.29 | **73.33** |
| Heart (Swiss) | 88.59 | **88.72** |
| Hepatitis | **80.58** | 79.88 |
| Horse-Colic | **76.75** | 74.77 |
| Image Segmentation | 92.38 | **93.33** |
| Ionosphere | **92.60** | 91.44 |
| Iris | 92.00 | **96.00** |
| Liver Disorders | 55.04 | **57.09** |
| Pima Indians Diabetes | **71.89** | 70.32 |
| Satellite Image | 87.06 | **89.33** * |
| Shuttle | 96.17 | **99.61** * |
| Sonar | 78.45 | **84.19** |
| Thyroid (Allbp) | 94.86 | **95.29** |
| Thyroid (Allhyper) | 96.93 | **97.50** |
| Thyroid (Allhypo) | 89.36 | **90.18** |
| Thyroid (Allrep) | 96.86 | **97.07** |
| Thyroid (Dis) | **98.29** | 98.00 |
| Thyroid (Hypothyroid) | 93.01 | **96.96** * |
| Thyroid (Sick) | 88.24 | **97.11** * |
| Thyroid (Sick-Euthyroid) | 88.82 | **94.40** * |
| Vehicle | 63.72 | **65.37** * |
| Vowel | 91.47 | **96.21** * |
| Wine | 94.38 | **97.22** * |
| **Average:** | 83.08 | **84.68** |

Table 7. Generalization of WVDM vs. DVDM.

On this set of databases, WVDM was an average of 1.6% more accurate than DVDM overall. WVDM had a higher average accuracy than DVDM on 23 out of the 34 databases, and was significantly higher on 9, while DVDM was only higher on 11 databases, and none of those differences were statistically significant.

Section 6 provides further comparisons of WVDM with other distance functions, including IVDM.

## 6. Empirical Comparisons and Analysis of Distance Functions

This section compares the distance functions discussed in this paper. A nearest neighbor classifier was implemented using each of six different distance functions: Euclidean (normalized by standard deviation) and HOEM as discussed in Section 2; HVDM as discussed in Section 3; DVDM and IVDM as discussed in Section 4; and WVDM as discussed in Section 5. Figure 12 summarizes the definition of each distance function.

| All functions use the same overall distance function: | $D(\boldsymbol{x},\boldsymbol{y}) = \sqrt{\sum_{a=1}^{m} d_a(x_a, y_a)^2}$ | | |
|---|---|---|---|
| Distance Function | Definition of $d_a(x_a, y_a)$ for each attribute type: | | |
| | Continuous | Linear Discrete | Nominal |
| Euclidean | $\dfrac{x_a - y_a}{\sigma_a}$ | $\longleftarrow$ | $\dfrac{x_a - y_a}{\sigma_a}$ |
| HOEM | $\dfrac{x_a - y_a}{range_a}$ | $\longleftarrow$ | 0 if $x_a = y_a$<br>1 if $x_a \neq y_a$ |
| **HVDM** | $\dfrac{x_a - y_a}{4\sigma_a}$ | $\longleftarrow$ | $\sqrt{vdm_a(x_a, y_a)}$ |
| DVDM | $vdm_a(disc_a(x_a), disc_a(y_a))$ | $\longrightarrow$ | $\sqrt{vdm_a(x_a, y_a)}$ |
| **IVDM** | $ivdm_a(x_a, y_a)$<br>Interpolate probabilities from range midpoints. | $\longrightarrow$ | $\sqrt{vdm_a(x_a, y_a)}$ |
| **WVDM** | $wvdm_a(x_a, y_a)$<br>Interpolate probabilities from adjacent values. | $\longrightarrow$ | $\sqrt{vdm_a(x_a, y_a)}$ |
| where $range_a = max_a - min_a$, and $vdm_a(x, y) = \sum_{c=1}^{C} \left| P_{a,x,c} - P_{a,y,c} \right|^2$ | | | |

Figure 12. Summary of distance function definitions.

Each distance function was tested on 48 datasets from the UCI machine learning databases,

again using 10-fold cross-validation. The average accuracy over all 10 trials is reported for each test in Table 8. The highest accuracy achieved for each dataset is shown in bold. The names of the three new distance functions presented in this paper (HVDM, IVDM and WVDM) are also shown in bold to identify them.

Table 8 also lists the number of instances in each database ("#Inst."), and the number of continuous ("Con"), integer ("Int", i.e., linear discrete), and nominal ("Nom") input attributes.

| | Distance Function | | | | | | # of inputs | | |
| Database | Euclid | HOEM | **HVDM** | DVDM | **IVDM** | **WVDM** | #Inst. | Con | Int | Nom |
|---|---|---|---|---|---|---|---|---|---|---|
| Annealing | 94.99 | 94.61 | 94.61 | 94.99 | **96.11** | 95.87 | 798 | 6 | 3 | 29 |
| Audiology | 60.50 | 72.00 | **77.50** | 77.50 | **77.50** | **77.50** | 200 | 0 | 0 | 69 |
| Audiology (test) | 41.67 | 75.00 | **78.33** | **78.33** | **78.33** | **78.33** | 26 | 0 | 0 | 69 |
| Australian | 80.58 | 81.16 | 81.45 | **83.04** | 80.58 | 82.46 | 690 | 6 | 0 | 8 |
| Breast Cancer | 94.99 | 95.28 | 94.99 | **95.57** | **95.57** | **95.57** | 699 | 0 | 9 | 0 |
| Bridges | 58.64 | 53.73 | 59.64 | 56.73 | **60.55** | 56.64 | 108 | 1 | 3 | 7 |
| Credit Screening | 78.99 | 81.01 | 80.87 | 80.14 | 80.14 | **81.45** | 690 | 6 | 0 | 9 |
| Echocardiogram | 94.82 | 94.82 | 94.82 | **100.00** | **100.00** | 98.57 | 132 | 7 | 0 | 2 |
| Flag | 48.95 | 48.84 | 55.82 | **58.76** | 57.66 | 58.74 | 194 | 3 | 7 | 18 |
| Glass | **72.36** | 70.52 | **72.36** | 56.06 | 70.54 | 71.49 | 214 | 9 | 0 | 0 |
| Heart Disease | 72.22 | 75.56 | 78.52 | 80.37 | 81.85 | **82.96** | 270 | 5 | 2 | 6 |
| Heart (Cleveland) | 73.94 | 74.96 | 76.56 | 79.86 | 78.90 | **80.23** | 303 | 5 | 2 | 6 |
| Heart (Hungarian) | 73.45 | 74.47 | 76.85 | **81.30** | 80.98 | 79.26 | 294 | 5 | 2 | 6 |
| Heart (Long-Beach-Va) | **71.50** | 71.00 | 65.50 | 71.00 | 66.00 | 68.00 | 200 | 5 | 2 | 6 |
| Heart (More) | 72.09 | 71.90 | 72.09 | 72.29 | **73.33** | **73.33** | 1541 | 5 | 2 | 6 |
| Heart (Swiss) | **93.53** | 91.86 | 89.49 | 88.59 | 87.88 | 88.72 | 123 | 5 | 2 | 6 |
| Hepatitis | 77.50 | 77.50 | 76.67 | 80.58 | **82.58** | 79.88 | 155 | 6 | 0 | 13 |
| Horse-Colic | 65.77 | 60.82 | 60.53 | 76.75 | **76.78** | 74.77 | 301 | 7 | 0 | 16 |
| House-Votes-84 | 93.12 | 93.12 | **95.17** | **95.17** | **95.17** | **95.17** | 435 | 0 | 0 | 16 |
| Image Segmentation | 92.86 | **93.57** | 92.86 | 92.38 | 92.86 | 93.33 | 420 | 18 | 0 | 1 |
| Ionosphere | 86.32 | 86.33 | 86.32 | **92.60** | 91.17 | 91.44 | 351 | 34 | 0 | 0 |
| Iris | 94.67 | 95.33 | 94.67 | 92.00 | 94.67 | **96.00** | 150 | 4 | 0 | 0 |
| LED+17 noise | 42.90 | 42.90 | **60.70** | **60.70** | **60.70** | **60.70** | 10000 | 0 | 0 | 24 |
| LED | **57.20** | **57.20** | 56.40 | 56.40 | 56.40 | 56.40 | 1000 | 0 | 0 | 7 |
| Liver Disorders | 62.92 | **63.47** | 62.92 | 55.04 | 58.23 | 57.09 | 345 | 6 | 0 | 0 |
| Monks-1 | **77.08** | 69.43 | 68.09 | 68.09 | 68.09 | 68.09 | 432 | 0 | 0 | 6 |
| Monks-2 | 59.04 | 54.65 | **97.50** | **97.50** | **97.50** | **97.50** | 432 | 0 | 0 | 6 |
| Monks-3 | 87.26 | 78.49 | **100.00** | **100.00** | **100.00** | **100.00** | 432 | 0 | 0 | 6 |
| Mushroom | **100.00** | **100.00** | **100.00** | **100.00** | **100.00** | **100.00** | 8124 | 0 | 1 | 21 |
| Pima Indians Diabetes | 71.09 | 70.31 | 71.09 | **71.89** | 69.28 | 70.32 | 768 | 8 | 0 | 0 |
| Promoters | 73.73 | 82.09 | **92.36** | **92.36** | **92.36** | **92.36** | 106 | 0 | 0 | 57 |
| Satellite Image | 90.21 | **90.24** | 90.21 | 87.06 | 89.79 | 89.33 | 4435 | 36 | 0 | 0 |
| Shuttle | **99.78** | **99.78** | **99.78** | 96.17 | 99.77 | 99.61 | 9253 | 9 | 0 | 0 |
| Sonar | **87.02** | 86.60 | **87.02** | 78.45 | 84.17 | 84.19 | 208 | 60 | 0 | 0 |
| Soybean (Large) | 87.26 | 89.20 | 90.88 | **92.18** | **92.18** | **92.18** | 307 | 0 | 6 | 29 |
| Soybean (Small) | **100.00** | **100.00** | **100.00** | **100.00** | **100.00** | **100.00** | 47 | 0 | 6 | 29 |
| Thyroid (Allbp) | 94.89 | 94.89 | 95.00 | 94.86 | **95.32** | 95.29 | 2800 | 6 | 0 | 22 |
| Thyroid (Allhyper) | 97.00 | 97.00 | 96.86 | 96.93 | **97.86** | 97.50 | 2800 | 6 | 0 | 22 |
| Thyroid (Allhypo) | 90.39 | 90.39 | 90.29 | 89.36 | **96.07** | 90.18 | 2800 | 6 | 0 | 22 |
| Thyroid (Allrep) | 96.14 | 96.14 | 96.11 | 96.86 | **98.43** | 97.07 | 2800 | 6 | 0 | 22 |
| Thyroid (Dis) | 98.21 | 98.21 | 98.21 | **98.29** | 98.04 | 98.00 | 2800 | 6 | 0 | 22 |
| Thyroid (Hypothyroid) | 93.42 | 93.42 | 93.36 | 93.01 | **98.07** | 96.96 | 3163 | 7 | 0 | 18 |
| Thyroid (Sick-Euthyroid) | 68.23 | 68.23 | 68.23 | 88.24 | **95.07** | 94.40 | 3163 | 7 | 0 | 18 |
| Thyroid (Sick) | 86.93 | 86.89 | 86.61 | 88.82 | 96.86 | **97.11** | 2800 | 6 | 0 | 22 |
| Vehicle | **70.93** | 70.22 | **70.93** | 63.72 | 69.27 | 65.37 | 846 | 18 | 0 | 0 |
| Vowel | **99.24** | 98.86 | **99.24** | 91.47 | 97.53 | 96.21 | 528 | 10 | 0 | 0 |
| Wine | 95.46 | 95.46 | 95.46 | 94.38 | **97.78** | 97.22 | 178 | 13 | 0 | 0 |
| Zoo | 97.78 | 94.44 | **98.89** | **98.89** | **98.89** | **98.89** | 90 | 0 | 0 | 16 |
| Average: | 80.78 | 81.29 | 83.79 | 84.06 | **85.56** | 85.24 | | | | |

Table 8. Summary of Generalization Accuracy

On this set of 48 datasets, the three new distance functions (HVDM, IVDM and WVDM) did substantially better than Euclidean distance or HOEM. IVDM had the highest average accuracy (85.56%) and was almost 5% higher on average than Euclidean distance (80.78%), indicating that it is a more robust distance function on these datasets, especially those with nominal attributes. WVDM was only slightly lower than IVDM with 85.24% accuracy. Somewhat surprisingly, DVDM was slightly higher than HVDM on these datasets, even though it uses discretization instead of a linear distance on continuous attributes. All four of the VDM-based distance functions outperformed Euclidean distance and HOEM.

Out of the 48 datasets, Euclidean distance had the highest accuracy 11 times; HOEM was highest 7 times; HVDM, 14; DVDM, 19; IVDM, 25 and WVDM, 18.

For datasets with no continuous attributes, all four of the VDM-based distance functions (HVDM, DVDM, IVDM and WVDM) are equivalent. On such datasets, the VDM-based distance functions achieve an average accuracy of 86.6% compared to 78.8% for HOEM and 76.6% for Euclidean, indicating a substantial superiority on such problems.

For datasets with no nominal attributes, Euclidean and HVDM are equivalent, and all the distance functions perform about the same on average except for DVDM, which averages about 4% less than the others, indicating the detrimental effects of discretization. Euclidean and HOEM have similar definitions for applications without any nominal attributes, except that Euclidean is normalized by standard deviation while HOEM is normalized by the range of each attribute. It is interesting that the average accuracy over these datasets is slightly higher for Euclidean than HOEM, indicating that the standard deviation may provide better normalization on these datasets. However, the difference is small (less than 1%), and these datasets do not contain many outliers, so the difference is probably negligible in this case.

One disadvantage with scaling attributes by the standard deviation is that attributes which almost always have the same value (e.g., a boolean attribute that is almost always 0) will be given a large weight—not due to scale, but because of the relative frequencies of the attribute values. A related problem can occur in HVDM. If there is a very skewed class distribution (i.e., there are many more instances of some classes than others), then the $P$ values will be quite small for some classes and quite large for others, and in either case the difference $|P_{a,x,c} - P_{a,y,c}|$ will be correspondingly small, and thus nominal attributes will get very little weight when compared to linear attributes. This phenomenon was noted by Ting (1994, 1996), where he recognized such problems on the *hypothyroid* dataset. Future research will address these normalization problems and look for automated solutions. Fortunately, DVDM, IVDM and WVDM do not suffer from either problem, because all attributes are scaled by the same amount in such cases, which may in part account for their success over HVDM in the above experiments.

For datasets with both nominal and continuous attributes, HVDM is slightly higher than Euclidean distance on these datasets, which is in turn slightly higher than HOEM, indicating that the overlap metric may not be much of an improvement on heterogeneous databases. DVDM, IVDM and WVDM are all higher than Euclidean distance on such datasets, with IVDM again in the lead.

### 6.1. Effects of Sparse Data

Distance functions that use VDM require some statistics to determine distance. We therefore hypothesized that generalization accuracy might be lower for VDM-based distance functions than for Euclidean distance or HOEM when there was very little data available, and that VDM-based functions would increase in accuracy more slowly than the others as more instances were made available, until a sufficient number of instances allowed a reasonable sample size to determine good probability values.

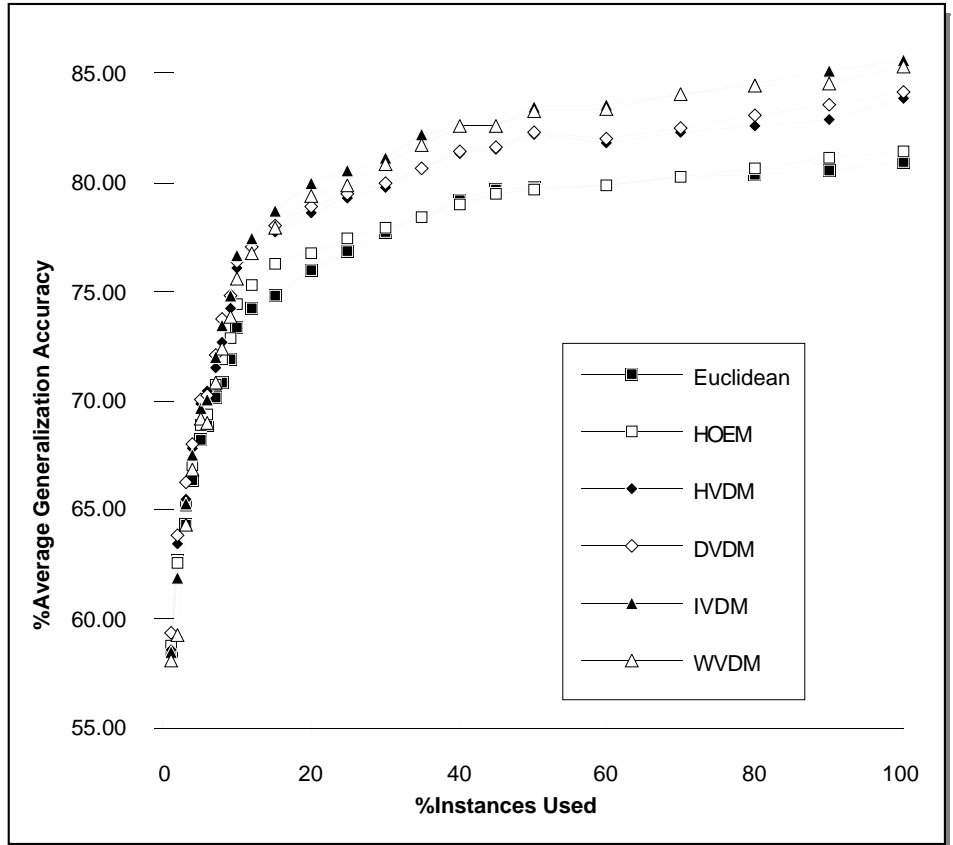

Figure 13. Average accuracy as the amount of data increases.

To test this hypothesis, the experiments used to obtain the results shown in Table 8 were repeated using only part of the available training data. Figure 13 shows how the generalization accuracy on the test set improves as the percentage of available training instances used for learning and generalization is increased from 1% to 100%. The generalization accuracy values shown are the averages over all 48 of the datasets in Table 8.

Surprisingly, the VDM-based distance functions increased in accuracy as fast or faster than Euclidean and HOEM even when there was very little data available. It may be that when there is very little data available, the random positioning of the sample data in the input space has a greater detrimental affect on accuracy than does the error in statistical sampling for VDM-based functions.

It is interesting to note from Figure 13 that the six distance functions seem to pair up into three distinct pairs. The interpolated VDM-based distance functions (IVDM and WVDM) maintain the highest accuracy, the other two VDM-based functions are next, and the functions based only on linear and overlap distance remain lowest from very early in the graph.

## 6.2. Efficiency Considerations

This section considers the storage requirements, learning speed, and generalization speed of each of the algorithms presented in this paper.

### 6.2.1. STORAGE

All of the above distance functions must store the entire training set, requiring O($nm$) storage, where $n$ is the number of instances in the training set and $m$ is the number of input attributes in the application, unless some instance pruning technique is used. For the Euclidean and HOEM functions, this is all that is necessary, but even this amount of storage can be restrictive as $n$ grows large.

For HVDM, DVDM, and IVDM, the probabilities $P_{a,x,c}$ for all $m$ attributes (only discrete attributes for HVDM) must be stored, requiring O($mvC$) storage, where $v$ is the average number of attribute values for the discrete (or discretized) attributes and $C$ is the number of output classes in the application. It is possible to instead store an array $D_{a,x,y} = vdm_a(x,y)$ for HVDM and DVDM, but the storage would be O($mv^2$), which is only a savings when $C < v$.

For WVDM, $C$ probability values must be stored for each continuous attribute value, resulting in O($nmC$) storage which is typically much larger than O($mvC$) because $n$ is usually much larger than $v$ (and cannot be less). It is also necessary to store a list of (pointers to) instances for each attribute, requiring an additional O($mn$) storage. Thus the total storage for WVDM is O(($C$+2)$nm$) = O($Cnm$).

| Distance Function | Storage | Learning Time | Generalization Time |
|---|---|---|---|
| Euclidean | O($mn$) | O($mn$) | O($mn$) |
| HOEM | O($mn$) | O($mn$) | O($mn$) |
| **HVDM** | O($mn$+$mvC$) | O($mn$+$mvC$) | O($mnC$) or O($mn$) |
| DVDM | O($mn$+$mvC$) | O($mn$+$mvC$) | O($mnC$) or O($mn$) |
| **IVDM** | O($mn$+$mvC$) | O($mn$+$mvC$) | O($mnC$) or O($mn$) |
| **WVDM** | O($Cmn$) | O($mn$log$n$+$mvC$) | O($mnC$) |

Table 9. Summary of efficiency for six distance metrics.

Table 9 summarizes the storage requirements of each system. WVDM is the only one of these distance functions that requires significantly more storage than the others. For most applications, $n$ is the critical factor, and all of these distance functions could be used in conjunction with instance pruning techniques to reduce storage requirements. See Section 7 for a list of several techniques to reduce the number of instances retained in the training set for subsequent generalization.

### 6.2.2. LEARNING SPEED

It takes $nm$ time to read in a training set. It takes an additional 2$nm$ time to find the standard deviation of the attributes for Euclidean distance, or just $nm$ time to find the ranges for HOEM.

Computing VDM statistics for HVDM, DVDM and IVDM takes $mn$+$mvC$ time, which is approximately O($mn$). Computing WVDM statistics takes $mn$log$n$+$mnC$ time, which is approximately O($mn$log$n$).

In general, the learning time is quite acceptable for all of these distance functions.

### 6.2.3. GENERALIZATION SPEED

Assuming that each distance function must compare a new input vector to all training instances, Euclidean and HOEM take O($mn$) time. HVDM, IVDM and DVDM take O($mnC$) (unless $D_{a,x,y}$ has been stored instead of $P_{a,x,c}$ for HVDM, in which case the search is done in O($mn$) time). WVDM takes O($\log n + mnC$) = O($mnC$) time.

Though $m$ and $C$ are typically fairly small, the generalization process can require a significant amount of time and/or computational resources as $n$ grows large. Techniques such as $k$-d trees (Deng & Moore, 1995; Wess, Althoff & Derwand, 1993; Sproull, 1991) and *projection* (Papadimitriou & Bentley, 1980) can reduce the time required to locate nearest neighbors from the training set, though such algorithms may require modification to handle both continuous and nominal attributes. Pruning techniques used to reduce storage (as in Section 6.2.1) will also reduce the number of instances that must be searched for generalization.

## 7. Related Work

Distance functions are used in a variety of fields, including instance-based learning, neural networks, statistics, pattern recognition, and cognitive psychology (see Section 1 for references). Section 2 lists several commonly-used distance functions involving numeric attributes.

Normalization is often desirable when using a linear distance function such as Euclidean distance so that some attributes do not arbitrarily get more weight than others. Dividing by the range or standard deviation to normalize numerical attributes is common practice. Turney (1993; Turney & Halasz, 1993) investigated *contextual normalization*, in which the standard deviation and mean used for normalization of continuous attributes depend on the context in which the input vector was obtained. In this paper we do not attempt to use contextual normalization, but instead use simpler methods of normalizing continuous attributes, and then focus on how to normalize appropriately between continuous and nominal attributes.

The Value Distance Metric (VDM) was introduced by Stanfill & Waltz (1986). It uses attribute weights not used by the functions presented in this paper. The Modified Value Difference Metric (MVDM) (Cost & Salzberg, 1993; Rachlin et al., 1994) does not use attribute weights but instead uses instance weights. It is assumed that these systems use discretization (Lebowitz, 1985; Schlimmer, 1987) to handle continuous attributes.

Ventura (1995; Ventura & Martinez, 1995) explored a variety of discretization methods for use in systems that can use only discrete input attributes. He found that using discretization to preprocess data often degraded accuracy, and recommended that machine learning algorithms be designed to handle continuous attributes directly.

Ting (1994, 1996) used several different discretization techniques in conjunction with MVDM and IB1 (Aha, Kibler & Albert, 1991). His results showed *improved* generalization accuracy when using discretization. Discretization allowed his algorithm to use MVDM on all attributes instead of using a linear distance on continuous attributes, and thus avoided some of the normalization problems discussed above in Sections 3.1 and 3.2. In this paper, similar results can be seen in the slightly higher results of DVDM (which also discretizes continuous attributes and then uses VDM) when compared to HVDM (which uses linear distance on continuous attributes). In this paper, DVDM uses equal-width intervals for discretization, while Ting's algorithms make use of more advanced discretization techniques.

Domingos (1995) uses a heterogeneous distance function similar to HVDM in his RISE system, a hybrid rule and instance-based learning system. However, RISE uses a normalization scheme similar to "N1" in Sections 3.1 and 3.2, and does not square individual attribute distances.

Mohri & Tanaka (1994) use a statistical technique called Quantification Method II (QM2) to derive attribute weights, and present distance functions that can handle both nominal and continuous attributes. They transform nominal attributes with $m$ values into $m$ boolean attributes, only one of which is on at a time, so that weights for each attribute can actually correspond to individual attribute values in the original data.

Turney (1994) addresses cross-validation error and voting (i.e. using values of $k > 1$) in instance-based learning systems, and explores issues related to selecting the parameter $k$ (i.e., number of neighbors used to decide on classification). In this paper we use $k = 1$ in order to focus attention on the distance functions themselves, but accuracy would be improved on some applications by using $k > 1$.

IVDM and WVDM use nonparametric density estimation techniques (Tapia & Thompson, 1978) in determining values of $P$ for use in computing distances. Parzen windows (Parzen, 1962) and shifting histograms (Rosenblatt, 1956) are similar in concept to these techniques, especially to WVDM. These techniques often use gaussian kernels or other more advanced techniques instead of a fixed-sized sliding window. We have experimented with gaussian-weighted kernels as well but results were slightly worse than either WVDM or IVDM, perhaps because of increased overfitting.

This paper applies each distance function to the problem of classification, in which an input vector is mapped into a discrete output class. These distance functions could also be used in systems that perform *regression* (Atkeson, Moore & Schaal, 1996; Atkeson, 1989; Cleveland & Loader, 1994), in which the output is a real value, often interpolated from nearby points, as in kernel regression (Deng & Moore, 1995).

As mentioned in Section 6.2 and elsewhere, pruning techniques can be used to reduce the storage requirements of instance-based systems and improve classification speed. Several techniques have been introduced, including IB3 (Aha, Kibler & Albert, 1991; Aha, 1992), the condensed nearest neighbor rule (Hart, 1968), the reduced nearest neighbor rule (Gates, 1972), the selective nearest neighbor rule (Rittler et al., 1975), typical instance based learning algorithm (Zhang, 1992), prototype methods (Chang, 1974), hyperrectangle techniques (Salzberg, 1991; Wettschereck & Dietterich, 1995), rule-based techniques (Domingos, 1995), random mutation hill climbing (Skalak, 1994; Cameron-Jones, 1995) and others (Kibler & Aha, 1987; Tomek, 1976; Wilson, 1972).

## 8. Conclusions & Future Research Areas

There are many learning systems that depend on a reliable distance function to achieve accurate generalization. The Euclidean distance function and many other distance functions are inappropriate for nominal attributes, and the HOEM function throws away information and does not achieve much better accuracy than the Euclidean function itself.

The Value Difference Metric (VDM) was designed to provide an appropriate measure of distance between two nominal attribute values. However, current systems that use the VDM often discretize continuous data into discrete ranges, which causes a loss of information and often a corresponding loss in generalization accuracy.

This paper introduced three new distance functions. The Heterogeneous Value Difference Function (HVDM) uses Euclidean distance on linear attributes and VDM on nominal attributes, and uses appropriate normalization. The Interpolated Value Difference Metric (IVDM) and Windowed Value Difference Metric (WVDM) handle continuous attributes within the same paradigm as VDM. Both IVDM and WVDM provide classification accuracy which is higher on average than the discretized version of the algorithm (DVDM) on the datasets with continuous attributes that we examined, and they are both equivalent to DVDM on applications without any continuous attributes.

In our experiments on 48 datasets, IVDM and WVDM achieved higher average accuracy than HVDM, and also did better than DVDM, HOEM and Euclidean distance. IVDM was slightly more accurate than WVDM and requires less time and storage, and thus would seem to be the most desirable distance function on heterogeneous applications similar to those used in this paper. Properly normalized Euclidean distance achieves comparable generalization accuracy when there are no nominal attributes, so in such situations it is still an appropriate distance function.

The learning system used to obtain generalization accuracy results in this paper was a nearest neighbor classifier, but the HVDM, IVDM and WVDM distance functions can be used with a $k$-nearest neighbor classifier with $k > 1$ or incorporated into a wide variety of other systems to allow them to handle continuous values including instance-based learning algorithms (such as PEBLS), radial basis function networks, and other distance-based neural networks. These new distance metrics can also be used in such areas as statistics, cognitive psychology, pattern recognition and other areas where the distance between heterogeneous input vectors is of interest. These distance functions can also be used in conjunction with weighting schemes and other improvements that each system provides.

The new distance functions presented here show improved average generalization on the 48 datasets used in experimentation. It is hoped that these datasets are representative of the kinds of applications that we face in the real world, and that these new distance functions will continue to provide improved generalization accuracy in such cases.

Future research will look at determining under what conditions each distance function is appropriate for a particular application. We will also look closely at the problem at selecting the window width, and will look at the possibility of smoothing WVDM's probability landscape to avoid overfitting. The new distance functions will also be used in conjunction with a variety of weighting schemes to provide more robust generalization in the presence of noise and irrelevant attributes, as well as increase generalization accuracy on a wide variety of applications.